\pdfoutput=1

\documentclass[11pt]{article}

\usepackage[preprint]{acl}

\usepackage{times}
\usepackage{latexsym}

\usepackage[T1]{fontenc}

\usepackage[utf8]{inputenc}

\usepackage{microtype}

\usepackage{inconsolata}

\usepackage{graphicx}

\usepackage{multirow}
\usepackage{xcolor}

\usepackage{float}
\usepackage{subfigure}

\newcommand\dunderline[2][.4pt]{%
  \raisebox{-#1}{\underline{\raisebox{#1}{\smash{\underline{#2}}}}}}

\newcommand\blfootnote[1]{%
  \begingroup
  \renewcommand\thefootnote{}\footnote{#1}%
  \addtocounter{footnote}{-1}%
  \endgroup
}

\title{On the Way to LLM Personalization: Learning to Remember User Conversations}

\author{
 \textbf{Lucie Charlotte Magister\textsuperscript{1,*}} \qquad
 \textbf{Katherine Metcalf\textsuperscript{2}} \qquad
 \textbf{Yizhe Zhang\textsuperscript{2} \qquad
 \textbf{Maartje ter Hoeve\textsuperscript{2}}} \\
\\
 \textsuperscript{1}University of Cambridge \\
 \textsuperscript{2}Apple \\ \\
 \texttt{lcm67@cam.ac.uk}, \texttt{\{kmetcalf, yizhe\_zhang, m\_terhoeve\}@apple.com}
}

\begin{document}
\maketitle

\blfootnote{*Work done while interning at Apple.}

\begin{abstract}
Large Language Models (LLMs) have quickly become an invaluable assistant for a variety of tasks. However, their effectiveness is constrained by their ability to tailor responses to human preferences and behaviors via personalization. Prior work in LLM personalization has largely focused on style transfer or incorporating small factoids about the user, as knowledge injection remains an open challenge. In this paper, we explore injecting knowledge of prior conversations into LLMs to enable future work on less redundant, personalized conversations. We identify two real-world constraints: (1) conversations are sequential in time and must be treated as such during training, and (2) per-user personalization is only viable in parameter-efficient settings. To this aim, we propose PLUM, a pipeline performing data augmentation for up-sampling conversations as question-answer pairs, that are then used to finetune a low-rank adaptation adapter with a weighted cross entropy loss. Even in this first exploration of the problem, we perform competitively with baselines such as RAG, attaining an accuracy of $81.5\%$ across 100 conversations.
\end{abstract}

\begin{figure*}[ht!]
  \includegraphics[width=\linewidth]{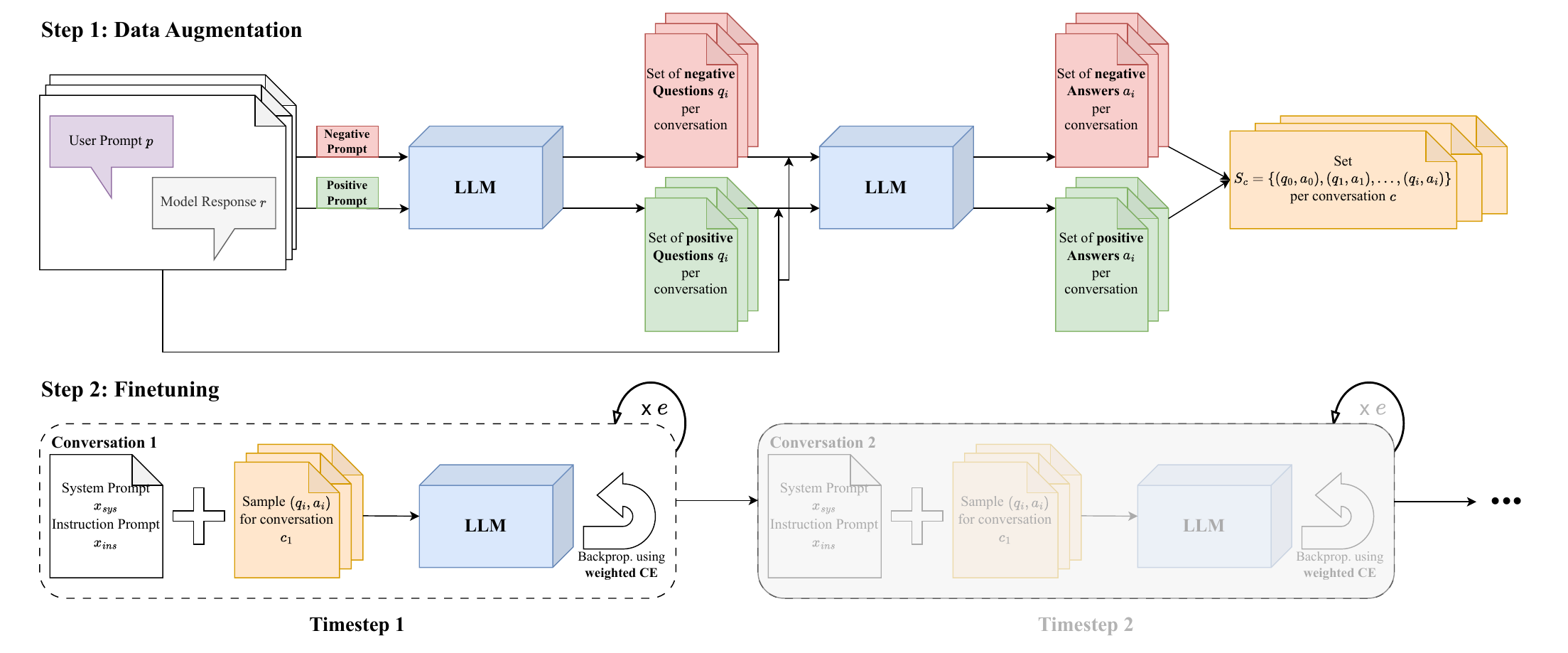}
  \caption{On overview of PLUM, a two-stage pipeline for injecting knowledge of prior user conversations into the LLM. The first step of the pipeline focuses on augmenting user conversations as positive and negative question-answer pairs about the conversation. These are then used in the finetuning step, where the LLM is trained on samples of a single conversation at a time for 10 epochs with a weighted cross entropy loss.}
  \label{visual_abstract}
\end{figure*}

\section{Introduction}
Large Language Models (LLMs) have quickly become a go-to resource for learning about new topics or assisting with a plethora of tasks. 
However, to fully unlock the models' capabilities, responses require personalization, tuning the model to the user's preferences and needs \citep{salemi2023lamp, zhuang2024hydra}. Prior work on LLM personalization has largely focused on adapting to the user's style and preferences via Reinforcement Learning from Human Feedback (RLFH) \citep{poddar2024personalizing, chen2024large, li2024personalized, maghakian2022personalized} or Parameter-Efficient Finetuning (PEFT) \citep{zhuang2024hydra, tan2024democratizinglargelanguagemodels}. To move beyond simple preference learning and small facts about the user, Retrieval Augmented Generation (RAG) methods have been used to integrate user profiles and conversation history into the model's generation process \citep{salemi2023lamp, mysore2023pearl, li2023teach, salemi2024optimization, zhang-etal-2024-llm-based}. However, RAG-based methods require maintaining external storage and picking a suitable number of documents to retrieve, with LLM performance having been shown to deteriorate with larger context windows \citep{vodrahalli2024michelangelo, zhang-etal-2024-bench}. Furthermore, \citet{wozniak2024personalized} show that personalized finetuning yields improved model reasoning over zero-shot prompting. This leads us to question whether parametric knowledge injection could yield a more streamlined approach to encoding knowledge of prior interactions with the user, opening new avenues for future LLM personalization.

Current works on knowledge injection mainly focus on adding fact-based knowledge to LLMs \citep{ovadia2023fine, fu2023revisiting, mecklenburg2024injecting} and note that knowledge injection remains an open challenge \citep{fu2023revisiting}. In this paper, we tackle the challenge of injecting knowledge of prior user conversations into LLMs via finetuning. Our efforts are in support of enabling future personalization research. Given the focus on user conversations, we identify and impose two important constraints: (1) finetuning must be parameter efficient and (2) conversations are sequential in nature and we do not want to store them like RAG-based methods. Note that we focus on inter-conversation rather than intra-conversation knowledge. We are interested in remembering the conversation holistically after it has occurred, as individual turns within a conversation can usually be injected into the context window. To the best of our knowledge, this is the first work considering to move beyond learning simple facts about the user to learning user conversations via finetuning. Specifically, we propose a \textit{\textbf{P}ipeline for \textbf{L}earning \textbf{U}ser Conversations in Large Language \textbf{M}odels} (PLUM), which extracts question-answer pairs from conversations to finetune a LLM with a Low-Rank Adaptation (LoRA) adapter \citep{hu2021lora} using a weighted cross entropy (CE) loss. In this initial exploration of the problem, PLUM already achieves an accuracy of $81.5\%$ across 100 conversations compared to $83.5\%$ with RAG-based baselines. Furthermore, we present an extensive set of ablations to guide and inspire future endeavors for the parametric personalization of LLMs, moving beyond simple RAG systems.

\section{Related Work}

\subsection{Personalization}
LLM personalization focuses on tuning model responses to the user's preferences and needs \citep{salemi2023lamp}. Existing works can broadly be split into three categories or a combination thereof: (1) prompting techniques, (2) RAG-based techniques and (3) RLHF and/or PEFT methods. Prompt-based techniques focus on encoding user preferences or conversation history in soft prompts \citep{hebert2024persoma, huang2024selective, shen2024pmg} or hard prompts \citep{mao2024reinforced, li2024learning}. Similarly, RAG-based techniques have been leveraged to add relevant information from a user's history to the LLM's context \citep{wu-etal-2021-personalized, salemi2023lamp, lu2023memochat, kumar2024longlamp, wang2024unims, neelakanteswara-etal-2024-rags, salemi2024optimization, zerhoudi2024personarag, sun2024knowledge, huang2024learning}, varying mostly in the type of information and the manner in which it is stored. However, a common challenge of these techniques is the reliance on prompt tuning and the selection of relevant data points \citep{zhuang2024hydra}. RLHF \citep{NEURIPS2020_1f89885d, jang2023personalized, cheng2023everyone, park2024principled, poddar2024personalizing, li2024personalized} and PEFT-based methods \citep{tan2024democratizinglargelanguagemodels, zhuang2024hydra} alleviate these issues by directly encoding user information in model parameters, however, they are usually limited to preferences or simple facts. In contrast, our work focuses on teaching the LLM to remember user conversations, which may span multiple facts and preferences. Specifically, we aim to provide an alternative to RAG by exploring injecting knowledge of previous user conversations.

\subsection{Knowledge Injection}
Knowledge injection focuses on adding new knowledge to the LLM after the initial training phase. Methods vary from prompt-based techniques \citep{chen2022knowprompt} to incorporating external knowledge sources via RAG \citep{song2016two, fan2020enhanced, lewis2020retrieval, martino2023knowledge, zhou2024cognitive, sun2024knowledge} to finetuning adapters \citep{wang2021k, ovadia2023fine, fu2023revisiting, mecklenburg2024injecting}. We take inspiration from \citet{mecklenburg2024injecting}, who explore how to augment data to directly incorporate it into the LLM via PEFT. However, our work diverges in the application to user conversation data with the aim of enabling future personalization research. Moreover, knowledge of user conversations can be seen as a special type of knowledge, as the contents are not completely foreign to the model but grounded in its knowledge \citep{gekhman-etal-2024-fine}.

\section{PLUM}
We propose PLUM, a two-stage pipeline for injecting knowledge of prior user conversations into LLMs. The first stage encompasses augmenting the user conversation data. The second stage focuses on PEFT with a custom loss function. We refer the reader to Figure \ref{visual_abstract} for a visual overview.

\subsection{Data Augmentation}
\label{data_gen_method}
We define a user conversation as a set of turns between the user and model, starting with an original user prompt. For simplicity, we will stick to single turn conversations, however, the proposed method can easily be extended to multi-turn. Recall that our goal is to enable future personalization by remembering user conversations, making single turn conversations a natural starting point. Given a user conversation $c$ defined as the tuple $(p, r)$, where $p$ is the user prompt and $r$ the LLM's response, we will use the same LLM to generate a set of question-answer pairs about the conversation $c$. This set of question-answer pairs can be denoted as $S_{c} = \{(q_0, a_0), (q_1, a_1), ..., (q_i, a_i)\}$, where $q_i$ is a question about the conversation and $a_i$ the corresponding answer to the question. To generate $S_{c}$, we provide the LLM with the original conversation and prompt it to ask as many questions as reasonable about the conversation. Note, that we do not specify a specific number of questions to be generated, as some conversations are more data-rich than others, i.e., for some conversations it is easier to generate many questions than for others.

We generate two types of questions using few-shot prompting for guidance. The first type are open-ended questions such as `What did we discuss about ...?', while the second are focused on eliciting a clear `yes' or `no' response, such as `Did we discuss ...?'. We then provide the original conversation and the individual questions about the conversation to the model for answering. Besides positive question-answer pairs, we also want to generate negative pairs. These are questions asking about something not covered by the conversation, eliciting a `no' response. This is to reinforce the knowledge boundary of the LLM and prevent hallucination, as we observe that without negative samples the LLM will default to always positively answering questions. To generate negative samples, we ask the LLM to pose questions adjacent to the topic of the conversation, to increase precision and not clash with other samples. We propose maintaining a balance between positive and negative samples so that the model does not err in one direction. Appendix \ref{gen_org_conv} and \ref{q_a_pair_gen} document our prompts.

\subsection{Parameter-Efficient Finetuning} 
We impose two design constraints for the injection of conversation history. First, we note that conversations are sequential in nature, which means that we should finish finetuning on one conversation before moving on to the next. This allows for discarding the conversation after all of its samples have been iterated over, which stands in contrast to RAG-based techniques. However, it also poses the challenge of overcoming catastrophic forgetting \citep{luo2023empirical}. Second, we note that finetuning all model parameters per user is infeasible, therefore, we use a PEFT method.

Given these constraints, we propose finetuning a LoRA adapter conversation by conversation. We propose a LoRA adapter based on its robust performance in the previous knowledge injection work of \citet{mecklenburg2024injecting}. To finetune the LoRA adapter, we use teacher forcing for more stable training \citep{williams1989learning}. In teacher forcing the model is provided with the true previous tokens rather than the predicted ones. Each training example $x_i$ consists of four key elements:
\begin{equation}
    x_i = x_{sys} + x_{ins} + a_i + q_i
\end{equation}

The first two elements are an optional system prompt $x_{sys}$ and an instruction prompt $x_{ins}$. These two elements are consistent across all training examples, while the final two elements are a question $q_i$ and corresponding answer $a_i$ sampled from the set $S_c$ for the conversation $c$. We refer the reader to Appendix \ref{model_finetuning_prompts} for the full prompt. 

In addition to this specific data setup, we empirically derive a custom loss based on the CE loss \citep{hinton2006fast}. We propose scaling up the CE loss on the question and answer tokens. We can rewrite the standard CE loss as a weighted version in the following way:
\begin{equation}
    L = H(P, Q_{(x_{sys}, x_{ins})}) + \lambda H(P, Q_{(x_{q_i}, x_{a_i})}),
\end{equation}
where $H$ is the CE loss measuring the difference between the true distribution of the data $P$ and the model's predicted distribution of the data $Q$. We compute the standard CE loss over the tokens $x_{sys}$ and $x_{ins}$. In contrast, we scale the CE loss over the tokens $q_i$ and $a_i$ of a training example by $\lambda$, specifically $\lambda = 10$. We empirically derive this loss after observing that the standard CE loss quickly diminishes with the model only having learned $x_{sys}$, $x_{ins}$ and the overall structure of the prompt well. We find that weighting $q_i$ and $a_i$ focuses the model on the actual question and answer, rather than just matching $x_{sys}$ and $x_{ins}$. We refer the reader to Section \ref{loss_ablations} for a discussion of ablations.

Based on our exploration, we specifically propose finetuning a LoRA adapter of rank $r=16$ and scaling parameter $\alpha=64$, that attaches to all linear layers in the LLM. We also suggest training on each conversation for $e=10$ epochs. To clarify, all data samples for a conversation are shown to the model for $10$ epochs before moving on to the next conversation. We also found a batch size of $b=8$ to work best, as well as maintaining a balance of positive and negative question-answer pairs. We further elaborate on these suggestions in Section \ref{results}.

\section{Experimental Setup}

\subsection{Data}
We rely on the OpenAssistant Conversations dataset \citep{NEURIPS2023_949f0f8f} for the initial human-generated prompt. We select 100 of these prompts as starting points for the conversations. We limit our initial exploration to 100 conversations to allow for more control during the data generation and because the user may also forget about prior conversations \citep{wixted1991form}. The selected starting prompts are focused on knowledge transfer, rather than completing a specific task, as we believe task-related queries are of more temporary nature. We then generate a response to complete the single-turn conversations. Recall, that we use these conversations to generate two types of question-answer pairs: positive and negative pairs, as shown in Figure \ref{visual_abstract}. We filter these pairs, checking that the questions and answers align with the expected format and directionality of the answer. We refer the reader to Appendix \ref{gen_org_conv} and \ref{q_a_pair_gen} for further details on prompting and filtering. After filtering, we have 3726 positive and negative question-answer pair samples across 100 conversations. We manually spot check the generated data to verify its quality. We withhold two questions per conversation for the test dataset. While the train dataset contains open-ended and closed questions, the test dataset only contains a positive and negative closed question per conversation that can be answered with `yes' and `no', respectively. We select `yes' and `no' types of questions for evaluation, as they have a clear target. We also train and test the LLM's performance on question-answer pairs not generated by itself but a larger LLM, to evaluate the reliance on in-distribution data and question-answer formulations.

\subsection{Model}
For our study of injecting conversation history into LLMs, we focus on Llama 3 8B Instruct \citep{dubey2024llama}. We choose this model because of its high performance on a variety of tasks given its reasonable size. As mentioned previously, we also employ a larger model to generate a second train and test dataset. For this, we use Llama 3 70B Instruct \citep{dubey2024llama} to generate an alternative version of the train and test dataset, that does not directly align with the model's distribution due to capacity and training data differences. This is further motivated by the work of \citet{hong2024scale}, who find that model scale matters for capturing the structure of language. We refer the reader to Appendix \ref{training_details} for further details on finetuning.

\subsection{Metrics}
As previously mentioned, we maintain a holdout dataset of questions that can be answered with a simple `yes' or `no'. We use these questions to evaluate the overall accuracy of the model after being finetuned on all conversations in the dataset. We also track the accuracy over time, which is an average of the model accuracy on all seen conversations for a given finetuning step. For example, after the model has seen the samples for $n$ conversations, it is evaluated on the corresponding evaluation questions for these $n$ conversations. Evaluating the accuracy over time provides an insight into the model's confidence in its answer and potential for catastrophic forgetting.

We also evaluate model accuracy on a suite of five common benchmarking tasks to ensure PLUM does not deteriorate the base performance of the model. We evaluate the base model and a PLUM-finetuned version on: (1) Measuring Massive Multitask Language Understanding (MMLU) \citep{hendrycks2020measuring}, (2) HellaSwag \citep{zellers2019hellaswag}, (3) ARC \citep{Clark2018ThinkYH}, (4) PIQA \citep{Bisk2020} and (5) Social IQA (SiQA) \citep{sap2019social}. We use the Language Model Evaluation Harness framework \citep{eval-harness} to perform a 1-shot and 5-shot evaluation on these tasks.

\subsection{Baselines}
The focus of our paper resides on enabling future personalization research by injecting knowledge of prior user conversations via PEFT. To contextualize our results, we compare our method to a standard RAG baseline, following \citet{salemi2023lamp}. Specifically, we train three retriever models based on BM25  \citep{robertson1995okapi}, a strong term matching model. The first BM25 baseline \textit{Conv. RAG} is trained on the original conversation data. The second baseline \textit{Sum. RAG} is trained only on summaries of the original conversations, to mimic common setups in existing literature \citep{richardson2023integrating}. Lastly, the \textit{Q/A RAG} baseline is a BM25 model trained on the question-answer pairs used for LoRA finetuning for a more comparable setup. We test different settings for $k$, the number of documents to retrieve, ranging from $k=\{1, 2, 3\}$. We choose BM25, because of its enduring strong performance \citep{salemi2024optimization, izacardunsupervised}. We do not compare to neural-based retrievers, such as Contriever \citep{izacardunsupervised}, because we only focus on 100 conversations for this study, which a neural retriever may easily overfit on. Moreover, our data setup for some baselines is not compatible with the training of Contriever, which requires positive and negative sample pairs for the contrastive loss.

\section{Results}
\label{results}

\begin{table}[ht]
  \centering
    \begin{tabular}{llccc}
    \hline
    \multicolumn{2}{l}{\multirow{2}{*}{\bf Model Setup}} & \multicolumn{3}{c}{\bf Accuracy (\%)} \\
     \multicolumn{2}{l}{} & Yes & No & Overall \\ 
    \hline
     \multicolumn{2}{l}{\color{purple}PLUM} & \color{purple}$73.0$ & \color{purple}$77.0$ & \color{purple}$75.0$ \\ 
     \multicolumn{2}{l}{\color{purple}PLUM (w/ sys.)} & \color{purple}$71.0$ & \color{purple}$92.0$ & \color{purple}$\mathbf{81.5}$ \\
     \hline
     \parbox[t]{1mm}{\multirow{8}{*}{\rotatebox[origin=c]{90}{Epochs}}} & $e=1$ & $82.0$ & $30.0$ & $56.0$ \\
     & $e=1$ (w/ sys.) & $70.0$ & $46.0$ & $58.0$ \\
     & $e=5$ & $70.0$ & $78.0$ & $74.0$ \\
     & $e=5$ (w/ sys.) & $84.0$ & $57.0$ & $70.5$ \\
     & $e=15$ & $39.0$ & $91.0$ & $65.0$ \\
     & $e=15$ (w/ sys.) & $91.0$ & $52.0$ & $71.5$ \\
     & $e=20$ & $71.0$ & $55.0$ & $63.0$ \\
     & $e=20$ (w/ sys.) & $0.0$ & $0.0$ & $0.0$ \\
     \hline
     \parbox[t]{1mm}{\multirow{3}{*}{\rotatebox[origin=c]{90}{CE }}} & $e=1$ & $5.0$ & $98.0$ & $51.5$ \\
     & $e=10$ & $21.0$ & $13.0$ & $17.0$ \\
     & $e=20$ & $0.0$ & $0.0$ & $0.0$\\
     & w/ sys. & $38.0$ & $44.0$ & $41.0$ \\
     \hline
     \parbox[t]{1mm}{\multirow{3}{*}{\rotatebox[origin=c]{90}{Loss Var. }}} & $(q_i, a_i)$-only CE & $88.0$ & $25.0$ & $56.5$ \\
     & $(q_i, a_i)$-only & $57.0$ & $95.0$ & \underline{$76.0$} \\
     & $a_i$-only & $40.0$ & $90.0$ & $65.0$ \\
     & $a_i$-only (w/ sys.) & $88.0$ & $35.0$ & $61.5$ \\
     \hline
     \parbox[t]{1mm}{\multirow{3}{*}{\rotatebox[origin=c]{90}{Data}}} & 70B Model Gen. & $30.0$ & $80.0$ & $55.0$ \\
     & Upsampled Yes & $75.0$ & $76.0$ & $75.5$ \\
     & Upsampled No & $69.0$ & $83.0$ & \underline{$76.0$} \\
     \hline
    \end{tabular}
  \caption{Model accuracy on various ablations. The best and second best overall accuracy are in \textbf{bold} and \underline{underlined}.}
  \label{ablation_studies}
\end{table}

\subsection{Ablations on Remembering User Conversations}
\label{overall_acc}

We perform an ablation study on various elements of PLUM to evaluate the contributions of the different components. By dissecting the elements of our framework, we aim to provide insights into the underlying mechanisms driving its performance and inspire future research. We present our results in Table \ref{ablation_studies}. Our ablations focus on the impact of epochs, the design of the loss function and the data. We also ablate our method with and without a system prompt. We provide further ablations on the LoRA architecture, batch size and random seeds in Appendix \ref{further_ablations}.

\subsubsection{Impact of Epochs}

We ablate the number of epochs required to remember a conversation. Recall that epochs refers to the number of times the model sees the training examples for a conversation before moving on to the next. We find that increasing or decreasing the number of epochs from $e=10$ deteriorates accuracy. Notably, for all settings except $e=10$ the model overfits to positive samples. For example, at $e=1$ we observe an imbalance in accuracy of $82.0\%$ versus $30.0\%$ for `yes' and `no' samples, respectively. This could be related to batching, as we backpropagate on $b=8$ samples at a time. Batching multiple examples may cause oscillation between erring on the positive and negative side, indicating adapting the number of epochs per conversation as an interesting area for further exploration. Furthermore, we observe the accuracy dropping to $0.0\%$ for two settings with $e=20$, as the model begins to output incoherent sentences.

\subsubsection{Impact of the Loss}
\label{loss_ablations}

We compare our weighted CE loss against the standard CE loss. Recall that in our weighted CE loss, we scale the loss of the question and answer tokens by $\lambda = 10$. We achieve an accuracy of $75.0\%$ without and $81.5\%$ with the system prompt. In comparison, simply training using the standard CE loss yields an accuracy of $17.0\%$ and $41.0\%$ without and with the system prompt, respectively. This highlights the significance of our design choice of up-weighting the loss on the question-answer section of the input prompt. If we only train on the question-answer pairs, this means we drop the system and instruction prompt, scaling the loss still achieves a significant improvement at an accuracy of $76.0\%$ compared to $56.5\%$. Lastly, we also examine whether only scaling the loss on the answer tokens is sufficient. We find that this only yields an overall accuracy of $65.0\%$ without and $61.5\%$ with the system prompt.

\begin{table}[ht]
\setlength{\tabcolsep}{4.5pt}
  \centering
    \begin{tabular}{llcc}
    \hline
    \multicolumn{2}{l}{\multirow{2}{*}{\bf Model Setup}} & \multicolumn{2}{c}{\bf Accuracy over Time (\%)} \\
     \multicolumn{2}{l}{} & Yes & No \\ 
    \hline
     \multicolumn{2}{l}{\color{purple}PLUM} & \color{purple}$68.2 \pm 17.9$ & \color{purple}$79.2 \pm 17.3$ \\ 
     \multicolumn{2}{l}{\color{purple}PLUM (w/ sys.)} & \color{purple}$79.7 \pm 18.3$ & \color{purple}$59.2 \pm 28.1$ \\
     \hline
     \parbox[t]{1mm}{\multirow{8}{*}{\rotatebox[origin=c]{90}{Epochs}}} & $e=1$ & $58.7 \pm 29.4$ & $52.2 \pm 29.9$ \\
     & $e=1$ (w/ sys.) & $45.7 \pm 25.7$ & $66.3 \pm 22.4$ \\
     & $e=5$ & $60.6 \pm 23.4$ & $70.4 \pm 26.8$ \\
     & $e=5$ (w/ sys.) & $81.8 \pm 23.4$ & $42.2 \pm 28.4$ \\
     & $e=15$ & $55.6 \pm 22.7$ & $79.5 \pm 15.2$ \\
     & $e=15$ (w/ sys.) & $82.9 \pm 19.3$ & $51.4 \pm 25.0$ \\
     & $e=20$ & $46.8 \pm 20.4$ & $79.4 \pm 13.5$ \\
     & $e=20$ (w/ sys.) & $30.8 \pm 39.0$ & $30.5 \pm 36.8$ \\
     \hline
     \parbox[t]{1.5mm}{\multirow{4}{*}{\rotatebox[origin=c]{90}{CE}}} & $e=1$ & $1.9 \pm 3.0$ & $98.9 \pm 1.7$ \\
     & $e=10$ & $54.7 \pm 35.0$ & $41.9 \pm 35.2$ \\
     & $e=20$ & $26.7 \pm 37.0$ & $22.3 \pm 32.4$ \\
     & w/ sys. & $54.9 \pm 34.1$ & $53.5 \pm 32.3$ \\
     \hline
     \parbox[t]{1mm}{\multirow{4}{*}{\rotatebox[origin=c]{90}{Loss Var.}}} & $(q_i, a_i)$-only CE & $63.1 \pm 33.4$ & $55.1 \pm 35.7$ \\
     & $(q_i, a_i)$-only & $59.4 \pm 17.7$ & $85.3 \pm 18.0$ \\
     & $a_i$-only & $46.9 \pm 21.6$ & $75.5 \pm 22.4$ \\
     & $a_i$-only (w/ sys.) & $82.5 \pm 21.6$ & $35.0 \pm 22.7$ \\
     \hline
     \parbox[t]{1mm}{\multirow{3}{*}{\rotatebox[origin=c]{90}{Data}}} & 70B Model Gen. & $39.1 \pm 27.4$ & $84.7 \pm 16.6$ \\
     & Upsampled `Yes' & $64.3 \pm 21.3$ & $70.9 \pm 24.5$ \\
     & Upsampled `No' & $67.3 \pm 20.5$ & $70.0 \pm 23.2$ \\
     \hline
    \end{tabular}
  \caption{Model accuracy \textbf{over time} (including standard deviation) on `yes' and `no' questions for various ablations.}
  \label{ablation_studies_over_time}
\end{table}

\subsubsection{Impact of Data}
We ablate our assumptions on the data balance between positive and negative samples, as well as staying within the model's distribution. We find that increasing the number of positive or negative samples by $25\%$ per conversation deteriorates results. Specifically, the model's accuracy to respond with `yes' or `no' increases for the respective cases, but decreases for the opposite. This indicates that maintaining a balance between samples is the most beneficial. Lastly, we also verify whether it is beneficial to stay within the model's distribution by training on samples not generated by the model itself. When training Llama 3 8B Instruct on data generated by its 70B counterpart, accuracy significantly deteriorates to only $30.0\%$ on `yes' questions. This can potentially be explained by training on data generated by the model itself reinforcing the knowledge and only requiring to store whether it was discussed previously. In contrast, the different wording and potentially divergent knowledge generated by the 70B model is not as simple to learn. We provide further results in Appendix \ref{large_model_split}.

\begin{table*}[ht]
  \centering
    \begin{tabular}{lcccccc}
    \hline
    \multirow{2}{*}{\bf Baselines} & \multicolumn{2}{c}{\bf Llama 3 8B Instruct} & \multicolumn{2}{c}{\bf PLUM} & \multicolumn{2}{c}{\bf PLUM (w/ sys.)} \\
     & \multicolumn{1}{c}{1-Shot} & \multicolumn{1}{c}{5-Shot} & \multicolumn{1}{c}{1-Shot} & \multicolumn{1}{c}{5-Shot} & \multicolumn{1}{c}{1-Shot} & \multicolumn{1}{c}{5-Shot} \\
     \hline
    MMLU & $64.2 \pm 0.4$ & $65.7 \pm 0.4$ &  $63.0 \pm 0.4$ & $64.9 \pm 0.4$ & $63.2 \pm 0.4$ & $65.3 \pm 0.4$ \\
    HellaSwag & $57.5 \pm 0.5$ & $58.4 \pm 0.5$ & $56.9 \pm 0.5$ & $57.8 \pm 0.5$ & $56.6 \pm 0.5$ & $57.2 \pm 0.5$ \\
    ARC-Easy & $83.9 \pm 0.8$ & $85.2 \pm 0.7$ &  $82.7 \pm 0.8$ & $83.6 \pm 0.8$ & $83.2 \pm 0.8$ & $83.5 \pm 0.8$ \\
    ARC-Cha. & $55.4 \pm 1.5$ & $57.4 \pm 1.4$ & $54.0 \pm 1.5$ & $56.1 \pm  1.5$ & $54.1 \pm 1.5$ & $55.5 \pm 1.5$ \\
    PiQA & $79.7 \pm 0.9$ & $80.6 \pm 0.9$  & $79.0 \pm 1.0$ & $80.1 \pm 0.9$ & $78.7 \pm 1.0$ & $80.0 \pm 0.9$ \\
    SiQA & $53.8 \pm 1.1$  & $56.8 \pm 1.1$ & $54.7 \pm 1.1$ & $57.9 \pm 1.1$ & $54.7 \pm 1.1$ & $56.7 \pm 1.1$ \\
    \hline
    \end{tabular}
  \caption{Model performance on a selection of benchmarking tasks before and after finetuning on user conversations. While we generally observe a slight deterioration in accuracy, performance remains within a reasonable range.}
  \label{benchmarking_tasks}
\end{table*}

\begin{figure}[t]
  \includegraphics[width=\columnwidth]{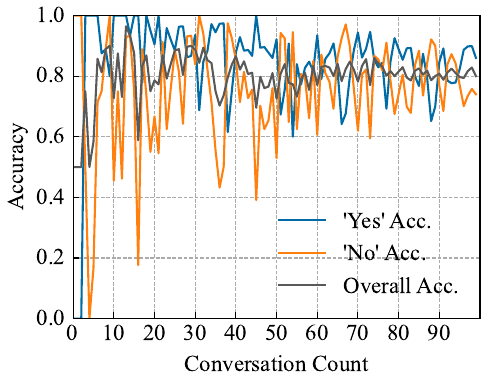}
  \caption{Accuracy \textbf{over time} for PLUM with the system prompt.}
  \label{accuracy_time_plot}
  \vspace{-0.5cm}
\end{figure}

\begin{figure*}[t]
    \centering
    \textbf{PLUM (w/ sys.)} \\
    \subfigure[Seen `Yes' Samples]{\includegraphics[width=0.23\textwidth]{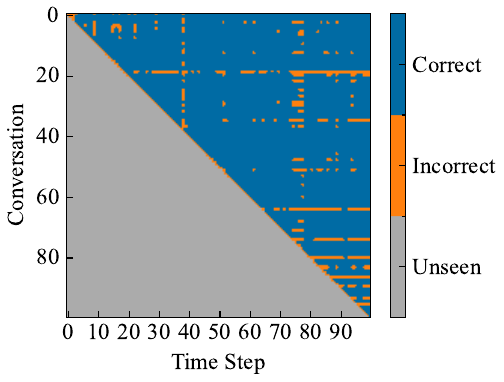}
    \label{plum_seen_yes}}    
    \subfigure[Seen `No' Samples]{\includegraphics[width=0.23\textwidth]{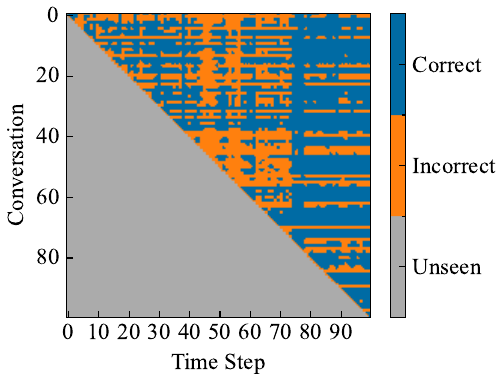}
    \label{plum_seen_no}}
    \subfigure[Unseen `Yes' Samples]{\includegraphics[width=0.23\textwidth]{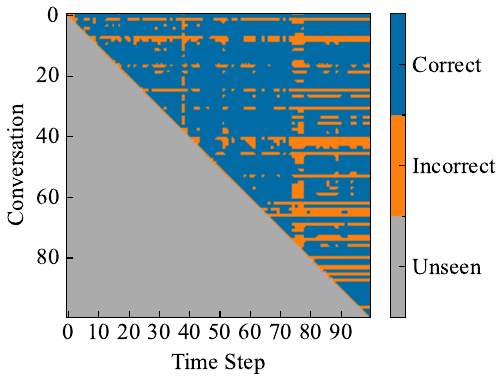}
    \label{plum_unseen_yes}}
    \subfigure[Unseen `No' Samples]{\includegraphics[width=0.23\textwidth]{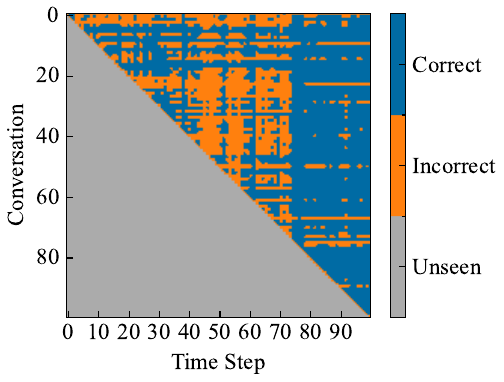}
    \label{plum_unseen_no}}

    \textbf{CE (e=10)} \\
    \subfigure[Seen `Yes' Samples]
    {\includegraphics[width=0.23\textwidth]{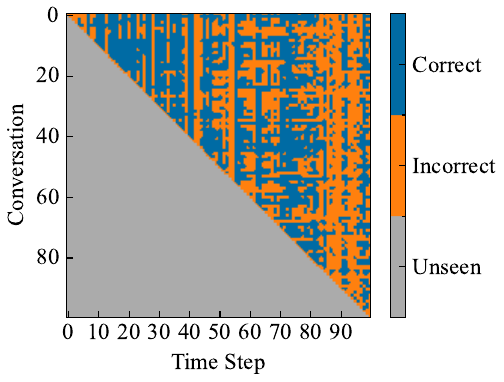}
    \label{ce_seen_yes}}
    \subfigure[Seen `No' Samples]{\includegraphics[width=0.23\textwidth]{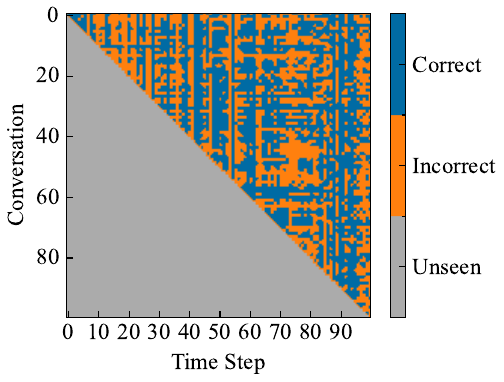}
    \label{ce_seen_no}}
    \subfigure[Unseen `Yes' Samples]{\includegraphics[width=0.23\textwidth]{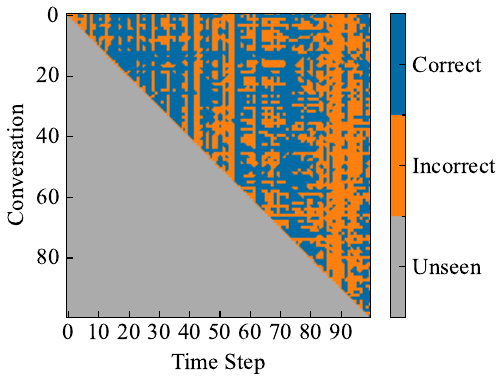}
    \label{ce_unseen_yes}}
    \subfigure[Unseen `No' Samples]{\includegraphics[width=0.23\textwidth]{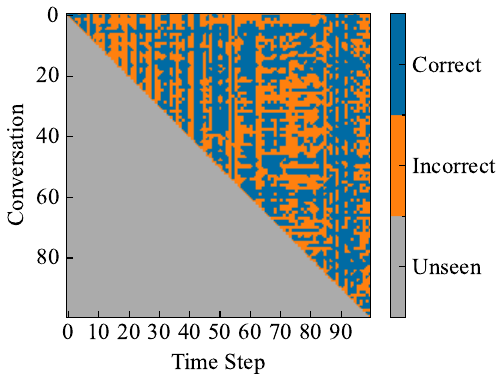}
    \label{ce_unseen_no}}
    \caption{Consistency plots visualizing whether the `yes'/`no' question was predicted correctly (blue) or incorrectly (orange) for a given time step. Here, a time step refers to the model having seen all samples of a conversation for the specified number of epochs. The lower left triangle of the plot is gray, as these conversations have not been seen yet.}
    \label{consistency_plots}
\end{figure*}

\subsection{Performance over Time}
We measure the model's accuracy over time to detect issues such as catastrophic forgetting. Table \ref{ablation_studies_over_time} summarizes the `yes' and `no' accuracy over time for various ablations. We observe similar trends as in our analysis of the overall model accuracy in Section \ref{overall_acc}, however, a noteworthy insight obtained from the accuracy over time is the standard deviation of the accuracy. For example, the average `yes' and `no' accuracy overtime for PLUM without a system prompt is $68.2\%$ and $79.2\%$ with a standard deviation of $\pm 17.9\%$ and $\pm 17.3\%$, respectively. This is quite a significant range, which indicates that the model oscillates between answering `yes' and `no'. Figure \ref{accuracy_time_plot} plots the accuracy as the model learns more conversations. We can observe that the model oscillates between erring on the `yes' or `no' side.

To further investigate this, we plot the predictions over time in Figure \ref{consistency_plots}. We call these plots \textit{consistency plots}, as ideally the upper triangle of each plot would be blue, meaning that the model consistently predicts the correct answer. We find that with PLUM, the LLM fairly consistently learns how to answer the questions for a given conversation, however, it fails to learn some conversations altogether, indicated by consistent streaks of orange. We observe no sign of catastrophic forgetting. Moreover, in Figure \ref{plum_unseen_no} it appears as if the LLM initially struggles to reply with `no', but then learns this, starting at around time step 70. This shift towards `no' is also recognizable in the corresponding `yes' Figure \ref{plum_unseen_yes}, as there are more streaks of orange starting at time step 70. This could indicate the the LLM learns to balance responding with `yes' and `no'. Nevertheless, we should observe that PLUM allows strides in the right direction, despite the oscillations. This is especially evident when contrasting the consistency plots of PLUM (Figures \ref{plum_seen_yes}-\ref{plum_unseen_no}) with those generated when using standard CE loss with $e=10$ (Figures \ref{ce_seen_yes}-\ref{ce_unseen_no}).

\subsection{Model Performance}
Table \ref{benchmarking_tasks} shows the 1-shot and 5-shot performance of Llama 3 8B Instruct and the PLUM-finetuned model on five common benchmarking tasks. We can observe a slight deterioration in accuracy across all tasks except SiQA. We observe a slight improvement in accuracy from $53.8\%$ to $54.7\%$ on the SiQA dataset, which may be attributed to training on prior conversations being somewhat related to reasoning about social settings. Overall, we can conclude that PLUM does not negatively impact performance on standard benchmarking tasks. 

\subsection{Comparison against Baselines}
Table \ref{rag_vs_finetuning} details the accuracy of PLUM against various baselines. PLUM with and without a system prompt achieves a competitive accuracy of $81.5\%$ and $75.0\%$, respectively. For context, we include the 0-shot performance of the Llama 3 8B Instruct model and a perfect recall version of RAG, where we automatically inject the correct conversation or summary into the prompt. The 0-shot performance of the model is $50.0\%$, as the model defaults to responding with `no'. The highest accuracy of $89.5\%$ is achieved when providing the conversation to the LLM (\textit{Perfect Conv. Recall}). We can see this as the near-optimal performance of the LLM on the given data, but not necessarily an upper bound, as this is dependent on data quality. In contrast, the best performing RAG baseline is \textit{Conv. RAG} with $k=3$ at $86.5\%$. We observe an increase in accuracy as $k$ increases, as the BM25 model does not always return the correct conversation as the top one. The \textit{Sum. RAG} baseline achieves only $71.0\%$ accuracy, which can be attributed to the summaries potentially missing details needed to answer the questions. The \textit{Q/A RAG} baseline is the most comparable to PLUM, as it has access to the same data. It achieves an accuracy of $83.5\%$ for $k=3$. The performance of PLUM is just shy of this, highlighting injecting conversation history into LLMs as a promising avenue for future research on personalization.

\begin{table}[ht]
  \centering
  \begin{tabular}{lccc}
        \hline
        \multirow{2}{*}{\bf Method} & \multicolumn{3}{c}{\bf Accuracy (\%)} \\
        & Yes & No & Overall \\
        \hline
         0-shot Performance & $0.0$ & $100.0$ & $50.0$ \\
         \hline
         \textcolor{gray}{Perfect Conv. Recall} & \textcolor{gray}{100.0} & \textcolor{gray}{79.0} & \textcolor{gray}{89.5} \\
         \textcolor{gray}{Perfect Sum. Recall} & \textcolor{gray}{83.0} & \textcolor{gray}{78.0} & \textcolor{gray}{80.5} \\
         \hline
         Conv. RAG (k=1) & $86.0$ & $80.0$ & $83.0$ \\
         Conv. RAG (k=2) & $84.0$ & $84.0$ & $\underline{84.0}$ \\
         Conv. RAG (k=3) & $84.0$ & $89.0$ & $\mathbf{86.5}$ \\
         \hline
         Sum. RAG (k=1) & $56.0$ & $85.0$ & $70.5$ \\
         Sum. RAG (k=2) & $67.0$ & $81.0$ & $74.0$ \\
         Sum. RAG (k=3) & $68.0$ & $74.0$ & $71.0$ \\
         \hline
         Q/A RAG (k=1) & $66.0$ & $94.0$ & $80.0$ \\
         Q/A RAG (k=2) & $71.0$ & $94.0$ & $82.5$ \\
         Q/A RAG (k=3) & $73.0$ & $94.0$ & $83.5$ \\
         \hline
         \color{purple}PLUM & \color{purple}$73.0$ & \color{purple}$77.0$ & \color{purple}$75.0$ \\
         \color{purple}PLUM (w/ sys.) & \color{purple}$71.0$ & \color{purple}$92.0$ & \color{purple}$81.5$ \\
         \hline
    \end{tabular}
  \caption{Performance of RAG-based baselines versus PLUM, with the best and second best overall accuracy in \textbf{bold} and \underline{underlined}.}
  \label{rag_vs_finetuning}
  \vspace{-0.5cm}
\end{table}

\section{Discussion}

\subsection{Take-aways}
We show that PLUM allows LLMs to efficiently remember user conversations. We can identify three key takeaways from our experiments that allow this success. First, the number of times the training samples for a conversation are shown to the model is highly important. We found that $e=10$ provides the best trade-off between the accuracy of `yes' and `no' questions. A lower or higher number of epochs generally leads the model to err towards `yes'. More importantly, we found that the weighted CE loss is necessary for the successful recall of the question-answer pairs. In our experiments without the weighted loss or different configurations of weighting tokens, we observed significant model deterioration. We also found that a balance of positive and negative samples is important for reinforcement of the knowledge boundary. Providing more positive samples deteriorates the performance on negative ones and vice versa.

\subsection{Future Work}

Our key takeaways indicate potential for future research. An area for further study is tuning the number of epochs per conversation, as some conversations may be easier to learn than others. Other avenues for exploration include further experiments on the loss and data sampling strategies. Another area of research is improving the answer consistency of the model, as we observe large variations between time steps. Lastly, PLUM enables future research in personalization. For example, the injected knowledge of conversations can be used to reduce redundancy in responses or refine knowledge transfer between the LLM and user by slowly building on past interactions.

\section{Conclusion}
In this work, we explore injecting knowledge of prior user conversations into LLMs. We propose PLUM, a pipeline for finetuning a LoRA adapter on question-answer pairs about prior conversations, maintaining conversation order. Moreover, we propose a custom loss for improved results. While we do not outperform the RAG baseline, we achieve competitive results. Our results indicate directly injecting personalization data into LLMs as an interesting avenue for future research, such as reducing redundancy in subsequent conversations and potentially extending reasoning capabilities.

\section{Limitations}

Despite the contributions of this work, our work must be viewed in the context of a few limitations. It should be noted that all data used is in English and that we only verified PLUM on Llama 3 8B Instruct. Moreover, we limited our study to only 100 conversations, which can be seen as a reasonable but small subset of conversations over time. Lastly, we did not explore model performance on remembering conversations on the same or clashing topics.

\section{Ethical Considerations}

We must carefully examine the ethical implications of our work. Remembering user conversations may lead to personal information being stored in the parametric knowledge of a LLM, which an adversary may extract \citep{mattern-etal-2023-membership}. This applies to personalization in general. Moreover, personalization may provide identifying information about the user leading to biases in the generated text \citep{wang2023coffee, he2024cos, weissburg2024llms}. Furthermore, future work should consider how a model personalized with prior user conversations should talk to the user, as in \citet{liao2021should}. We urge the reader to carefully consider the aforementioned points in their work extending or using PLUM to handle the user's privacy with care.

\section*{Acknowledgements}
We are thankful to Stéphane Aroca-Ouellette, Moises Goldszmidt, Natalie Mackraz, Natalie Schluter, and Skyler Seto for their helpful discussions, comments, and thoughtful feedback.

\bibliography{references}

\appendix

\section{Generating the Original Conversation}
\label{gen_org_conv}

\subsection{Conversation Prompt}
\label{conv_prompts}
We use the OpenAssistant Dataset \citep{NEURIPS2023_949f0f8f} to sample 100 initial conversation prompts. These conversation prompts are human generated and in English. As we are looking for knowledge-based prompts, we hand-select 100 of these. 

\subsection{Conversation Response Generation}
\label{conv_resp_gen}
We use the following system prompt to generate a response to the initial starting prompt from the OpenAssistant Dataset \citep{NEURIPS2023_949f0f8f}:

\texttt{You are a helpful LLM answering questions concisely. Do not ramble or generate repetitive output.}

We then simply add the original conversation prompt. Examining the dataset, we found that some of the prompts are not formatted correctly so we apply simple corrections such as capitalizing the first word of a sentence and adding a question mark at the end of sentences.

\section{Question-Answer Pair Generation}
\label{q_a_pair_gen}

We generate question-answer pairs via a two step process. First we infer a set of questions about the conversation. Then we simply prompt the model to answer the question about the conversation, providing the conversation as context. In total, we generate 4 types of question-answer pairs:

\begin{enumerate}
    \item Positive Open-ended Questions
    \item Negative Open-ended Questions
    \item Positive Closed-ended Questions (`Yes' Questions)
    \item Negative Closed-ended Questions (`No' Questions)
\end{enumerate}

We generate open-ended questions about the conversation to elicit a summary style answer. We generate positive and negative questions of this type, where positive means that the topic was indeed discussed in the conversation, while negative means that the topic was not discussed and the model should politely express this. Similarly, we also generate closed-ended questions of this type. These are questions that should be answerable with a `yes' and `no'.

\subsection{4-shot Sample Conversations}

In order to elicit open and closed-ended questions from the model, we perform 4-shot prompting. We write four knowledge-based single-turn conversations for this. We phrase some initial starting prompts as questions and others as instructions (e.g. `Tell me about ...'). We ensure that the responses to the questions have enough content to write a varied set of positive and negative sample questions.

\subsection{4-shot Sample Questions}

For each of the four few-shot sample conversations, we then write 12 positive and negative, open-ended and closed questions for each of the conversations. Recall that closed questions are of the format `Did we discuss...?' to elicit a `yes' and `no' response. The open-ended questions begin with `What did we discuss about ...?' to elicit a longer, summary-style responses. The 4-shot prompting with example questions allows us to demonstrate to the model that we are looking for positive questions, questions that can directly be answered via the conversation, and negative questions, questions about information adjacent to the topic of discussion. Note that we define 12 questions of each style of question for each few-shot example to encourage the model to write a substantial number of questions. However, we do not directly prompt the model to generate this many questions later on, to make sure that the questions are of high quality. We discuss this further in Section \ref{question_gen}.

\subsection{Question Generation}
\label{question_gen}

\subsubsection{Positive Questions}
In order to elicit the model to generate similar positive questions for our sample conversations generated on the base of the OpenAssistant dataset \citep{NEURIPS2023_949f0f8f}, we prompt it in the following way with the 4-shot examples prepended. We define the system prompt $x_{sys}$ as:

\texttt{You are a helpful LLM specialized in inferring the topic of a conversation and writing questions about what was discussed about this topic in the conversation. Do not deviate from the topics and contents of the conversation. All questions must be answerable with `yes'. Only speak from the `we' perspective of `you and the user'. You must start your sentence with `Did we discuss...'. Always specify the topic you are referring to, avoiding ambiguity, so that the question makes sense without the conversation. Never say `the conversation'.}

We then add the following instruction prompt $x_{ins}$:

\texttt{From the user perspective, write as many questions as sensible about what was discussed in the following conversation. <START\_CONVERSATION> USER: \textcolor{orange}{\{user\_prompt\}} YOU: \textcolor{orange}{\{model\_response\}} <END\_CONVERSATION>}

As indicated above, we inject the user prompt and model response in place of \textcolor{orange}{\{user\_prompt\}} and \textcolor{orange}{\{model\_response\}}, respectively. The system prompt and instruction prompt for the other types of questions follow a similar structure. Note that we only ask the model to generate as many questions about the conversation as sensible. This is because some conversations may be more data-rich than others, which makes it easier to ask varied questions. We observed this behavior during prompt tuning. Please also note that we have removed line breaks for formatting purposes here.

\subsubsection{Negative Questions}
In order to elicit negative questions, we replace the system prompt $x_{sys}$ with the following:

\texttt{You are a helpful LLM specialized in inferring the topic of a conversation and writing 12 questions about closely related topics that were not covered in the conversation. Do not deviate from the overarching topic of the conversation. All questions must be answerable with `We did not discuss ...'. Only speak from the `we' perspective of `you and the user'. Start your sentence with `What did we discuss about...'. Always specify the topic you are referring to, avoiding ambiguity, so that the question makes sense without the conversation. Never say `the conversation'.}

Similarly, we replace the instruction prompt $x_{ins}$ with the following: 

\texttt{From the user perspective, write 12 questions about something that was not discussed in the following conversation. <START\_CONVERSATION> USER: \textcolor{orange}{\{user\_prompt\}} YOU: \textcolor{orange}{\{model\_response\}} <END\_CONVERSATION>}

Please note that we include line breaks were appropriate, but we have removed these for the formatting of the paper. Please also note that we specify 12 questions here, rather than just asking the model to generate as many questions as sensible. This is because negative questions, questions about topics adjacent to the conversation, are easier to generate. However, please also notice that we then balance the number of positive and negative samples per conversation by taking the smaller number of the two.

\subsection{Answer Generation}
\label{answer_gen}

After splitting the questions generated, we prompt the model to answer each question individually. We also provide the user conversation as reference for the positive questions. We do not provide the conversation for the negative ones, as the information is not needed and we can rely on the performance of the base model, which is to politely decline having knowledge of prior conversations.

We use the following positive system prompt $x_{sys}$ to answer the questions about the conversation: 

\texttt{You are a helpful LLM trained to answer questions about prior conversations you had. You are very detailed and summarize the whole conversation to answer the question. You do not include details that are not in the conversation.}

We then provide the conversation and the instruction prompt $x_{ins}$:

\texttt{---Conversation--- USER: \textcolor{orange}{\{user\_prompt\}} AGENT: \textcolor{orange}{\{model\_response\}} --Task-- Please answer the following question about whether you have discussed the indicated topic with the user. Question: \textcolor{orange}{\{conv\_question\}} Answer:}

The model then begins completing the prompt. As before, we have marked the corresponding sections to add in the user prompt, model response and generated conversation question. The negative instruction follows a similar wording, however, we do not provide the conversation as context there. This is because the question cannot be answered from the conversation and the model should politely decline. Please note that we have also removed line breaks here for formatting purposes.

\subsection{Data Filtering}

We perform some basic checks for filtering. Firstly, we check that the questions generated follow the expected structure, e.g. `Did we discuss ...?'. We also check that the generated answer yields the expected response, e.g. it contains `yes'. We also remove all duplicate questions.

\section{Prompts for Model Finetuning}
\label{model_finetuning_prompts}

We use the following system prompt $x_{sys}$ for model finetuning:

\texttt{You are a helpful LLM trained to answer questions about prior conversations you had. If you do not remember having discussed the topic, you state to the user that you do not remember having had this conversation.}

We then follow-up with the following instruction prompt $x_{ins}$:

\texttt{Please answer the following question about whether you have discussed the indicated topic.}

We then concatenate the question $q_i$ and answer $a_i$ in the following format:

\texttt{Question \textcolor{orange}{\{question\}} Answer: \textcolor{orange}{\{answer\}}}

We insert the question-answer pair samples for a conversation in place of \textcolor{orange}{\{question\}} and \textcolor{orange}{\{answer\}}, respectively. All prompts presented were selected by writing variants and manually examining the quality of the performance for a handful of conversations. Again, please note that we have removed line breaks in our prompts for formatting purposes.

\begin{table*}[ht]
  \centering
    \begin{tabular}{llccc|ccc}
    \hline
    \multicolumn{2}{l}{\multirow{2}{*}{\bf Model Setup}} & \multicolumn{3}{c|}{\bf Accuracy (\%)} & \multicolumn{3}{c}{\bf Accuracy over Time {(\%)}} \\
    \multicolumn{2}{l}{} & Yes & No & Overall & Yes & No & Overall \\ 
    \hline
     \multicolumn{2}{l}{PLUM} & $73.0$ & $77.0$ & $75.0$ & $68.2 \pm 17.9$ & $79.2 \pm 17.3$ & $\mathbf{73.7 \pm 6.9}$ \\ 
     \multicolumn{2}{l}{PLUM (w/ sys.)} & $71.0$ & $92.0$ & $\mathbf{81.5}$ & $79.7 \pm 18.3$ & $59.2 \pm 28.1$ & \dunderline{$69.5 \pm 9.3$}\\
     \hline
     \parbox[t]{1mm}{\multirow{8}{*}{\rotatebox[origin=c]{90}{Epochs}}} & $e=1$ & $82.0$ & $30.0$ & $56.0$ & $58.7 \pm 29.4$ & $52.2 \pm 29.9$ & $55.4 \pm 4.8$ \\
     & $e=1$ (w/ sys.) & $70.0$ & $46.0$ & $58.0$ & $45.7 \pm 25.7$ & $66.3 \pm 22.4$ & $56.0 \pm 3.8$ \\
     & $e=5$ & $70.0$ & $78.0$ & $74.0$ & $60.6 \pm 23.4$ & $70.4 \pm 26.8$ & $65.5 \pm 9.1$\\
     & $e=5$ (w/ sys.) & $84.0$ & $57.0$ & $70.5$ & $81.8 \pm 23.4$ & $42.2 \pm 28.4$ & $62.0 \pm 7.8$ \\
     & $e=15$ & $39.0$ & $91.0$ & $65.0$ & $55.6 \pm 22.7$ & $79.5 \pm 15.2$ & $67.5 \pm 7.6$\\
     & $e=15$ (w/ sys.) & $91.0$ & $52.0$ & $71.5$ & $82.9 \pm 19.3$ & $51.4 \pm 25.0$ & $67.2 \pm 7.6$ \\
     & $e=20$ & $71.0$ & $55.0$ & $63.0$ & $46.8 \pm 20.4$ & $79.4 \pm 13.5$ & $63.0 \pm 6.2$ \\
     & $e=20$ (w/ sys.) & $0.0$ & $0.0$ & $0.0$ & $30.8 \pm 39.0$ & $30.5 \pm 36.8$ & $30.6 \pm 33.6$ \\
     \hline
     \parbox[t]{1mm}{\multirow{3}{*}{\rotatebox[origin=c]{90}{Batch}}} & $b=1$ & $84.0$ & $40.0$ & $62.0$ & $44.9 \pm 30.3$ & $58.3 \pm 30.3$ & $51.6 \pm 15.4$\\
     & $b=16$ & $83.0$ & $74.0$ & \dunderline{$78.5$} & $67.8 \pm 24.7$ & $68.5 \pm 26.4$ & $68.2 \pm 9.5$ \\
     & $b=32$ & $72.0$ & $75.0$ & $73.5$ & $57.3 \pm 24.9$ & $72.8 \pm 25.1$ & $65.0 \pm 9.1$ \\
     \hline
     \parbox[t]{1mm}{\multirow{4}{*}{\rotatebox[origin=c]{90}{CE}}} & $e=1$ & $5.0$ & $98.0$ & $51.5$ & $1.9 \pm 3.0$ & $98.9 \pm 1.7$ & $50.4 \pm 0.7$ \\
     & $e=10$ & $21.0$ & $13.0$ & $17.0$ & $54.7 \pm 35.0$ & $41.9 \pm 35.2$ & $48.3 \pm 16.7$ \\
     & $e=20$& $0.0$ & $0.0$ & $0.0$ & $26.6 \pm 37.0$ & $22.3 \pm 32.4$ & $24.5 \pm 26.6$ \\
     & w/ sys. & $38.0$ & $44.0$ & $41.0$ & $54.9 \pm 34.1$ & $53.5 \pm 32.3$ & $54.2 \pm 10.1$\\
     \hline
     \parbox[t]{1mm}{\multirow{4}{*}{\rotatebox[origin=c]{90}{Loss Var.}}} & $(q_i, a_i)$-only CE & $88.0$ & $25.0$ & $56.5$ & $63.1 \pm 33.4$ & $55.1 \pm 35.7$ & $59.1 \pm 13.5$ \\
     & $(q_i, a_i)$-only & $57.0$ & $95.0$ & $76.0$ & $59.4 \pm 17.7$ & $85.3 \pm 18.0$ & $72.4 \pm 6.9$ \\
     & $a_i$-only & $40.0$ & $90.0$ & $65.0$ & $46.9 \pm 21.6$ & $75.5 \pm 22.4$ & $61.2 \pm 7.1$ \\
     & $a_i$-only (w/ sys.) & $88.0$ & $35.0$ & $61.5$ & $82.5 \pm 21.6$ & $35.0 \pm 22.7$ & $58.8 \pm 5.2$ \\
     \hline
     \parbox[t]{1mm}{\multirow{3}{*}{\rotatebox[origin=c]{90}{Data}}} & 70B Model Gen. & $30.0$ & $80.0$ & $55.0$ & $39.1 \pm 27.4$ & $84.7 \pm 16.6$ & $61.9 \pm 10.1$\\
     & Upsampled `Yes' & $75.0$ & $76.0$ & $75.5$ & $64.3 \pm 21.3$ & $70.9 \pm 24.5$ & $67.6 \pm 7.1$ \\
     & Upsampled `No' & $69.0$ & $83.0$ & $76.0$ & $67.3 \pm 20.5$ & $70.0 \pm 23.2$ & $68.7 \pm 7.9$ \\
     \hline
     \parbox[t]{1mm}{\multirow{4}{*}{\rotatebox[origin=c]{90}{LoRA}}} & Att.-only, $r=16, \alpha=64$ & $92.0$ & $34.0$ & $63.0$ & $75.1 \pm 26.1$ & $42.9 \pm 28.7$ & $59.0 \pm 7.2$ \\
     & Lin., $r=16, \alpha=32$ & $85.0$ & $65.0$ & $75.0$ & $69.5 \pm 24.1$ & $69.9 \pm 23.1$ & $69.7 \pm 8.1$ \\
     & Lin., $r=8, \alpha=64$ & $64.0$ & $89.0$ & $76.5$ & $54.4 \pm 22.5$ & $85.3 \pm 18.1$ & $69.8 \pm 8.1$ \\
     & Lin., $r=32, \alpha=64$ & $63.0$ & $90.0$ & $76.5$ & $66.3 \pm 20.2$ & $80.6 \pm 19.5$ & $73.5 \pm 7.8$ \\
     \hline
     \parbox[t]{1mm}{\multirow{2}{*}{\rotatebox[origin=c]{90}{Seed}}} & $seed=7$ & $62.0$ & $81.0$ & $71.5$ & $69.6 \pm 19.0$ & $68.5 \pm 22.1$ & $69.0 \pm 6.6$ \\
     & $seed=73$ & $78.0$ & $78.0$ & $78.0$ & $66.2 \pm 22.2$ & $78.0 \pm 22.4$ & $72.1 \pm 9.8$
     \\
     \hline
    \end{tabular}
  \caption{Model performance on various ablations. The best and second best overall accuracy and accuracy over time are in \textbf{bold} and \underline{underlined}. For completeness, we have included all ablations run.}
\end{table*}

\section{Finetuning}
\label{training_details}

We perform teacher-forcing for finetuning, using the Adam optimizer \citep{kingma2014adam} with a learning rate of 0.001. We set the random seed to $s = 42$. All other hyperparameters are described in the main text. We train our models in a distributed manner, with a maximum number of 8 NVIDIA A100 80 GB being used for an experiment.

\begin{table*}[ht]
  \centering
  \begin{tabular}{lccc|ccc}
        \hline
        \multirow{2}{*}{\bf Method} & \multicolumn{3}{c|}{\bf Test Split Accuracy} & \multicolumn{3}{c}{\bf Llama 70B Gen. Split Accuracy} \\
        & Yes & No & Overall & Yes & No & Overall \\
        \hline
         0-shot Performance & $0.0$ & $100.0$ & $50.0$ & $0.0$ & $100.0$ & $50.0$ \\
         \hline
         \textcolor{gray}{Perfect Conv. Recall} & \textcolor{gray}{100.0} & \textcolor{gray}{79.0} & \textcolor{gray}{89.5} & \textcolor{gray}{100.0} & \textcolor{gray}{97.0} & \textcolor{gray}{98.5} \\
         \textcolor{gray}{Perfect Sum. Recall} & \textcolor{gray}{83.0} & \textcolor{gray}{78.0} & \textcolor{gray}{80.5} & \textcolor{gray}{100.0} & 
         \textcolor{gray}{95.0} & 
         \textcolor{gray}{97.5} \\
         \hline
         Conv. RAG (k=1) & $86.0$ & $80.0$ & $83.0$ & $93.0$ & $95.0$ & $94.0$ \\
         Conv. RAG (k=2) & $84.0$ & $84.0$ & $\underline{84.0}$ & $92.0$ & $89.0$ & $90.5$ \\
         Conv. RAG (k=3) & $84.0$ & $89.0$ & $\mathbf{86.5}$ & $93.0$ & $92.0$ & $92.5$ \\
         \hline
         Sum. RAG (k=1) & $56.0$ & $85.0$ & $70.5$ & $86.0$ & $96.0$ & $91.0$ \\
         Sum. RAG (k=2) & $67.0$ & $81.0$ & $74.0$ & $90.0$ & $89.0$ & $89.5$ \\
         Sum. RAG (k=3) & $68.0$ & $74.0$ & $71.0$ & $95.0$ & $84.0$ & $89.5$ \\
         \hline
         Q/A RAG (k=1) & $66.0$ & $94.0$ & $80.0$ & $82.0$ & $96.0$ & $89.0$ \\
         Q/A RAG (k=2) & $71.0$ & $94.0$ & $82.5$ & $93.0$ & $96.0$ & $\underline{94.5}$ \\
         Q/A RAG (k=3) & $73.0$ & $94.0$ & $83.5$ & $94.0$ & $97.0$ & $\mathbf{95.5}$ \\
         \hline
         PLUM & $73.0$ & $77.0$ & $75.0$ & $83.0$ & $73.0$ & $78.0$ \\
         PLUM (w/ sys.) & $71.0$ & $92.0$ & $81.5$ & $89.0$ & $45.0$ & $67.0$ \\
         \hline
    \end{tabular}
  \caption{Performance of RAG-based baselines versus PLUM on the Llama 8B and 70B model generated data splits. The best and second best overall accuracy are in \textbf{bold} and \underline{underlined}.}
  \label{app_rag_vs_finetuning}
\end{table*}

\section{Further Ablations}
\label{further_ablations}

\subsection{LoRA Architecture}
To gain a better understanding of how to best configure the LoRA adapter, we vary the layers the LoRA adapter attaches to and the adapter size. We found that attaching to all linear layers in the model yields the best results, as $81.5\%$ accuracy, compared to only attaching to attention layers, yielding an accuracy of only $63.0\%$. We also vary the adapter size $r$, with $r=8$ and $r=32$ only achieving an accuracy of $76.5\%$. Overall, we find that a LoRA adapter attaching to all linear layers, with a size of  $r=16$ and $a=64$, performs best. 

\subsection{Batch Size}
To gain a better understanding of the effect of the batch size we replicate our setup with different batch sizes where $b=\{1, 16, 32\}$. Note that the batch size we use for PLUM is $b=8$. We observe that a batch size of $b=16$ yields slightly improved results at $78.5\%$ compared to a $b=8$ at $75.0\%$ accuracy. However, the oscillations between `yes' and `no' are smaller at $b=8$, indicated by the decreased gap between the `yes' and `no' accuracy, as well as the standard deviation for the accuracy across time.

\subsection{Reproducibility}

To ensure that our results are reproducible, we run PLUM (without a system prompt) on two further seeds ($s=7$ and $s=73$). We observe similar performance at an overall accuracy of $71.5\%$ and $78.0\%$. 

\section{Performance on Llama 3 70B Instruct Generated Split}
\label{large_model_split}

Figure \ref{app_rag_vs_finetuning} summarizes the results of PLUM in comparison to RAG on the dataset generated with Llama 3 70B Instruct model. We observe that RAG significantly benefits from the data generated by the larger model. For example, \textit{Q/A RAG} with $k=1$ improves from an accuracy of $80.0\%$ to $89.0\%$. In contrast, PLUM does not benefit as much from the data generated by the larger model, with the accuracy only increasing from $75.0\%$ to $78.0\%$ in the version without the system prompt. In the case of PLUM with a system prompt, accuracy even deteriorates. This could be due to the model having a harder time remembering user conversations outside of its own model distribution. Using a different model to generate conversations may cause the model to have to remember new knowledge as well as whether a topic has been discussed or not, which is a more challenging task. Therefore, these results should be seen in context of PLUM requiring finetuning versus RAG simply performing retrieval.

\end{document}